\documentclass{article}

\usepackage{microtype}
\usepackage{graphicx}
\usepackage{booktabs}
\usepackage{amsmath}
\usepackage{amssymb}
\usepackage{multirow}
\usepackage{array}
\usepackage{xspace}
\usepackage{tikz}
\usetikzlibrary{positioning,fit,arrows.meta,patterns,decorations.pathreplacing}
\usepackage{hyperref}

\usepackage[accepted]{mlsys2025}
\makeatletter
\renewcommand{\Notice@String}{Preprint.}
\makeatother
\hypersetup{pdfauthor={Vyom Sharma, Debajyoti Datta}}

\usepackage{iftex}
\ifXeTeX\usepackage{fontspec}\fi

\graphicspath{{figures/}}
\newcommand{\sys}{RaMP\xspace}
\newcommand{\x}{$\times$\xspace}

\mlsystitlerunning{Runtime-Aware Megakernel Polymorphism for Mixture-of-Experts}

\begin{document}

\twocolumn[
\mlsystitle{RaMP: Runtime-Aware Megakernel Polymorphism\texorpdfstring{\\}{ }for Mixture-of-Experts}

\begin{mlsysauthorlist}
\mlsysauthor{Vyom Sharma}{hai}
\mlsysauthor{Debajyoti Datta}{hai}
\end{mlsysauthorlist}
\mlsysaffiliation{hai}{Hippocratic AI}
\mlsyscorrespondingauthor{Vyom Sharma}{vyom@hippocraticai.com}

\mlsyskeywords{Mixture of Experts, Kernel Selection, Expert Parallelism, Tensor Parallelism, LLM Serving, GPU Kernels}

\vskip 0.3in

\begin{abstract}
LLM serving systems autotune their GPU kernels offline by timing candidate kernel configurations, like tile sizes, at a range
of batch sizes. This works when
the number of tokens determines a kernel's work, as in dense layers. However, we show that it does not hold for
Mixture-of-Experts (MoE) layers where the router decides how the tokens are split across experts
and GPUs. We find that the best kernel configuration depends on this split, and not
only on the number of tokens---under expert parallelism, choosing the configuration for each GPU's actual
token distribution is up to 1.45\x faster than the existing choice. The split is not known until the router runs, so it cannot be looked up in advance. Hence, we introduce \sys, which chooses the kernel configuration at run time from the routing split itself. \sys compiles top configurations into one polymorphic kernel, and a small cost model on the GPU picks among them in a few microseconds. On serving traffic its cost model was never fit on, the configurations \sys's cost model chooses from the kernel's full configuration space are up to \textbf{1.45\x} faster than those vLLM's offline autotuner picks for the same kernel on batches of 128 tokens or more (1.27\x over all batches), and within 3.5\% of an exhaustive search that times every configuration. \sys's kernel and choice together make the MoE layer up to 1.84\x faster than vLLM's Triton MoE on batches of 128 tokens or more. In vLLM,
under both tensor and expert parallelism, \sys lowers time per output token by up to 1.11\x over the fastest stock MoE backend.
\end{abstract}

]

\printAffiliationsAndNotice{}

\section{Introduction}
\label{sec:intro}

\begin{figure}[!t]
\centering
\includegraphics[width=\columnwidth]{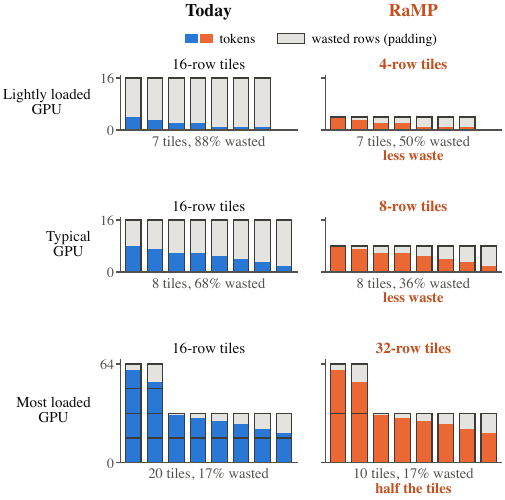}
\vskip -0.1in
\caption{\textbf{Before and after \sys}, on three GPUs of one batch (illustrative loads). Today every GPU runs the tile
size chosen from the batch size. \sys chooses per GPU: smaller tiles waste fewer rows where experts
receive few tokens, and larger tiles halve the tiles where they receive many.}
\label{fig:teaser}
\end{figure}

Many of the largest recent language models are Mixture-of-Experts (MoE)
models~\citep{shazeer2017outrageously,deepseekv3,qwen3moe,mixtral}. Each MoE layer holds many
experts, and a router sends each token to a few of them, so a megakernel computes all experts'
matrix multiplications on whatever tokens each one received~\citep{vllm,megablocks}. Large MoEs are
served with tensor parallelism (TP), which splits every expert across the GPUs, or with expert
parallelism (EP), which places whole experts on different GPUs~\citep{megatron,gshard,deepspeedmoe}.

Like any GPU kernel, the MoE kernel is fastest when its configuration, such as its tile sizes, fits
its work. For a dense layer, the work is set by the batch. For an MoE layer, it is set by the batch
\emph{and} by how balanced the routing is: how evenly the router spreads the batch's tokens across
experts and, under EP, across GPUs. Both change throughout serving. Continuous
batching~\citep{orca} changes the batch on every forward pass, and a production server receives
requests from many domains, which the router treats differently. In traces from vLLM serving eight
MoE models, a few experts often receive several times their even share, and which experts these
are changes with the requests in the batch.

Serving systems nevertheless choose the configuration offline, from the batch size alone:
vLLM~\citep{vllm} tunes the kernel at a fixed set of batch sizes on uniformly random routing and
uses the nearest entry at run time. When routing is unbalanced, that entry is often the wrong one:
under EP, the best configuration for each GPU is up to 1.45\x faster than vLLM's tuning.

The core problem is to choose from each GPU's expert load in real time. A batch's routing exists only
after the router runs, on the GPU, just before the kernel must start. Each configuration is a
separately compiled kernel chosen before the layer runs, and timing candidates on every batch would
take far longer than the kernel itself. Systems that adapt MoE execution at run time change the
parallelism or the expert placement~\citep{tutel,eplb}, not the kernel configuration.

Therefore, we introduce \sys, which chooses the MoE kernel configuration on every GPU, for every
forward pass, from the routing that GPU actually receives (Figure~\ref{fig:teaser}). \sys compiles many configurations into
one polymorphic kernel that reads which one to run from GPU memory, so switching between them needs
no new launch and costs about 1\%.
A small cost model ranks the configurations from the work the GPU's tokens produce, and
the GPU picks the best in a few microseconds, so the choice adds almost no latency.

On serving traffic the cost model was never fit on, the configurations it chooses from the kernel's full configuration space are up to
1.45\x faster than those of vLLM's tuning for the same kernel on batches of 128 tokens or more
(up to 1.27\x over all batches), and within 3.5\% of exhaustive search, an oracle that times every
configuration, on every EP layout we test. Its kernel and choice together make the MoE layer up to 1.84\x faster than vLLM's Triton MoE on batches of 128 tokens or more. Its kernel is also the fastest MoE backend we measure: end to end in
vLLM, under both TP and EP, \sys lowers time per output token at every load we test, by up to
1.11\x over the fastest stock backend on DeepSeek-V3, Qwen3-30B, and OLMoE, with unchanged model
quality.

In summary, we make two contributions that rely on each other:
\begin{itemize}
\item A cost model, fit offline, that ranks kernel configurations from each GPU's current routing
  at run time, cheaply enough to run on the GPU on every forward pass; and
\item A polymorphic MoE megakernel that compiles a menu of configurations, drawn from 98--337
  per model, into one kernel and runs whichever one the GPU chooses, written in CuTe
  DSL~\citep{cutlass}.
\end{itemize}

\section{Background}
\label{sec:background}

\subsection{MoE Geometry During Inference}
\label{sec:moe}

An MoE layer replaces the dense MLP with $E$ experts, and a router sends each token to $k$ of
them~\citep{shazeer2017outrageously,gshard}. Each expert is a gated MLP: it multiplies the tokens
it receives by a $d \times 2h$ matrix for the gate and up projections and an $h \times d$ matrix for the down projection, where $d$ is the hidden size and
$h$ the expert's intermediate size. This geometry varies widely across models: Mixtral-8x22B has 8
experts with $h = 16{,}384$, while Qwen3.5-35B has 256 experts with $h = 512$.

How the model is spread over GPUs sets the geometry each GPU sees. On $G$ GPUs, tensor parallelism
(TP)~\citep{megatron} gives every GPU all $E$ experts, each cut to width $h/G$, and sums the
partial outputs with an all-reduce. Expert parallelism (EP)~\citep{gshard,deepspeedmoe} gives every
GPU $E/G$ whole experts and exchanges tokens with an all-to-all before and after the layer; it has
become a common way to serve the largest MoEs~\citep{deepseekv3,eplb}. We write EP8 for EP over 8
GPUs. Either way, each GPU holds a fixed set of experts $\mathcal{E}_g$ with fixed weight shapes.
What it does not know in advance is how many tokens each of them will receive.

\subsection{GPU Kernels on This Geometry}
\label{sec:kernel}

An MoE megakernel computes all of a GPU's experts in one launch~\citep{vllm,megablocks}. Like any
matrix multiplication on a GPU, it cuts its output into tiles. Each tile goes to a
thread block on one of the GPU's streaming multiprocessors (SMs; H200 has 132), which streams
slices of both inputs from memory into shared memory and accumulates the tile on the tensor
cores~\citep{cutlass}. The tile sizes decide how well this works. A larger tile reuses every value
it loads across more of the output, so it reads less memory per operation, but it needs more
shared memory and registers, fewer tiles fit on an SM at once, and a small product yields too few
tiles to keep every SM busy. Whether a kernel is limited by memory or by arithmetic depends on how
many output rows share each loaded weight~\citep{roofline}, so the best tile sizes change with the
shape of the product, and the wrong ones are costly. Across the eight models we study, the tile
sizes that are fastest on batches of fewer than 32 tokens run 1.7--6.1\x slower than the best ones
on batches of 512 tokens or more.

A \emph{configuration} $c$ fixes the rows per tile $b_c$, the tile width across the output, the
pipeline depth, and the order in which tiles launch. In an MoE kernel, the rows of each expert's
product are the $n_e$ tokens routed to expert $e$. The kernel cuts them into tiles of $b_c$ rows
and pads the last one: an expert with 5 tokens and $b_c = 16$ fills one tile and pads 11 rows
(Figure~\ref{fig:tiling}). On
GPU $g$, with load $\mathbf{n}_g = (n_e)_{e \in \mathcal{E}_g}$, configuration $c$ launches $T_c$
tiles, which run in $W_c$ waves:
\begin{equation}
T_c(\mathbf{n}_g) = q_c \sum_{e \in \mathcal{E}_g} \left\lceil \frac{n_e}{b_c} \right\rceil,
\qquad
W_c(\mathbf{n}_g) = \left\lceil \frac{T_c(\mathbf{n}_g)}{S_c} \right\rceil.
\label{eq:tiles}
\end{equation}
Here $q_c$ is the number of tiles across the output, which the tile width and the geometry fix,
and $S_c$ is the number of tiles that run at once: the number of SMs times the thread blocks of $c$
that fit on each. If $T_c$ is slightly more than $S_c$,
the last wave leaves most SMs idle. Small tiles pad less when experts receive few tokens; large
tiles need fewer waves. The geometry fixes $q_c$ and the weights each tile reads. The tokens decide
the rest.

\begin{figure}[t]
\centering
\begin{tikzpicture}[font=\scriptsize, x=0.8cm, y=0.8cm]
\definecolor{tok}{HTML}{2A78D6}\definecolor{tokl}{HTML}{B7D3F6}\definecolor{ink2}{HTML}{52514E}
\tikzset{pad/.style={pattern=north east lines, pattern color=black!35, draw=black!45},
         tile/.style={draw=black!55, line width=0.5pt}}

\begin{scope}[shift={(0,0)}]
  \node[anchor=west, font=\footnotesize\bfseries] at (-0.3,2.45) {(a) Tiles};
  \fill[tokl] (0,0.4) rectangle (2.0,1.6);
  \foreach \r in {0,...,5} \draw[white, line width=0.4pt] (0,0.4+0.2*\r) -- (2.0,0.4+0.2*\r);
  \fill[pad] (0,0.0) rectangle (2.0,0.4);
  \foreach \c in {0,0.5,1.0,1.5} {
    \draw[tile] (\c,0.8) rectangle ++(0.5,0.8);
    \draw[tile] (\c,0.0) rectangle ++(0.5,0.8);
  }
  \draw[tok, line width=1.1pt] (0,0.8) rectangle ++(0.5,0.8);
  \draw[|-|, ink2] (-0.15,0.8) -- node[left, text=ink2] {$b_c$} (-0.15,1.6);
  \draw[|-|, ink2] (0,1.75) -- node[above, text=ink2, yshift=-1pt] {output width} (2.0,1.75);
  \node[text=ink2] at (1.0,-0.22) {$q_c = 4$ tiles across};
  \node[text=ink2, anchor=west] at (2.05,0.2) {pad};
  \draw[|-|, ink2] (2.15,0.4) -- node[right, text=ink2] {$n_e$} (2.15,1.6);
\end{scope}

\begin{scope}[shift={(3.45,0)}]
  \node[anchor=west, font=\footnotesize\bfseries] at (-0.15,2.45) {(b) Padding};
  \fill[pad] (0,0.5) rectangle (0.5,0.8);
  \fill[tok] (0,0) rectangle (0.5,0.5);
  \draw[tile] (0,0) rectangle (0.5,0.8);
  \node[text=ink2] at (0.25,-0.22) {$b_c{=}8$};
  \fill[pad] (1.05,0.5) rectangle (1.55,1.6);
  \fill[tok] (1.05,0) rectangle (1.55,0.5);
  \draw[tile] (1.05,0) rectangle (1.55,1.6);
  \node[text=ink2] at (1.3,-0.22) {$b_c{=}16$};
  \node[text=ink2] at (0.78,1.95) {5 tokens};
\end{scope}

\begin{scope}[shift={(6.7,0)}]
  \node[anchor=west, font=\footnotesize\bfseries] at (-0.75,2.45) {(c) Waves};
  \foreach \c in {0,1,2,3} \fill[tok] (0.55*\c,1.0) rectangle ++(0.45,0.45);
  \fill[tok] (0,0.2) rectangle ++(0.45,0.45);
  \foreach \c in {1,2,3} \draw[black!40, dashed] (0.55*\c,0.2) rectangle ++(0.45,0.45);
  \node[text=ink2, anchor=east] at (-0.05,1.225) {wave 1};
  \node[text=ink2, anchor=east] at (-0.05,0.425) {wave 2};
  \draw[|-|, ink2] (0,1.6) -- node[above, text=ink2, yshift=-1pt] {$S_c$ tiles at a time} (2.1,1.6);
  \node[text=ink2] at (1.05,-0.22) {3 slots idle};
\end{scope}
\end{tikzpicture}
\vskip -0.05in
\caption{\textbf{A configuration on one expert.} (a) Tiles of $b_c$ rows, $q_c$ across. (b) The
last tile is padded. (c) Tiles run $S_c$ at a time.}
\label{fig:tiling}
\end{figure}

\subsection{Routing on This Geometry}
\label{sec:routingbg}

\begin{figure*}[t]
\centering
\input{figures/routing.tex}
\vskip -0.05in
\caption{\textbf{Routing 32 tokens to 8 experts on 4 GPUs.} Bars: tokens per expert; numbers:
tokens per GPU. (a) Every expert and GPU receives an even share. (b) Experts differ, GPUs do not.
(c) Experts and GPUs both differ.}
\label{fig:routing}
\end{figure*}

Serving systems run one forward pass at a time over all active requests, carrying one token for
every request that is generating output and, with chunked prefill, part of the prompt of newly
arrived requests~\citep{orca,vllm,sglang,sarathi}. The batch of $B$ tokens therefore changes on
every forward pass, and every MoE layer routes it differently (Figure~\ref{fig:routing}). The $n_e$
sum to $Bk$, so perfectly
balanced routing gives every expert $Bk/E$ tokens. Under TP, $\mathcal{E}_g$ holds every expert, so
every GPU computes the whole load. Under EP, each GPU computes only its share $\mathbf{n}_g$, and
the layer waits for the slowest GPU.

\begin{figure}[t]
\centering
\includegraphics[width=\columnwidth]{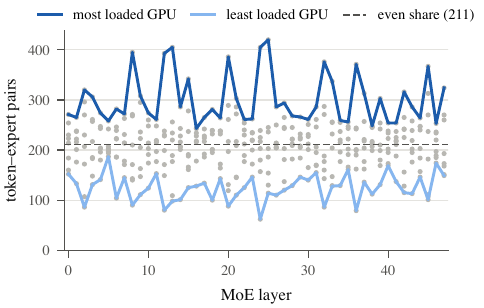}
\vskip -0.1in
\caption{\textbf{Per-GPU load in one forward pass.} Qwen3-30B on EP8, 211 tokens. Dots: each GPU's
token--expert pairs in each MoE layer; lines: the most and least loaded GPU; dashed: the even share,
$Bk/G$. A lookup by batch size sees 211 tokens on every GPU.}
\label{fig:onestep}
\end{figure}

Real routing is far from balanced. In the forward pass of Figure~\ref{fig:onestep}, Qwen3-30B
carries 211 tokens, and a lookup by batch size sees 211 on every GPU; on EP8, the most loaded GPU
receives up to twice the mean in some layers, and the least loaded under a third of it.

Routing matters to the kernel because it decides how the tokens fill tiles. Consider two GPUs that
each hold 16 experts and receive 128 tokens. On the first, routing is balanced and every expert
receives 8 tokens: 8-row tiles are exactly full, while 32-row tiles are three-quarters padding. On
the second, four experts receive 30 tokens each and eight others receive
one token each: 8-row tiles now
cut each busy expert into four tiles, and each tile reads that expert's weights again, while one
32-row tile per expert reads them once. Both GPUs have the same number of tokens, and Equation~\ref{eq:tiles}
favors a different configuration on each. Under EP this happens on every forward pass, on every
GPU, and the slowest GPU sets the layer's time.
Whether a given change in the load moves the best configuration, and by how much, depends on the
geometry and on the batch.

Real serving shows the same effect. In one recorded batch of 256 tokens of Qwen3-30B on EP8, the
GPUs' fastest configurations range from 6- to 32-row tiles and grow with the
tokens each GPU receives, while vLLM's tuning gives every GPU 24-row tiles. Across
seven models on EP8, the eight GPUs of a batch with 128 or more tokens need 4--6 different
configurations on average, and in at most 16\% of these batches is a single configuration best on
every GPU.

\subsection{Choosing a Configuration}
\label{sec:today}

MoE serving systems, like most GPU kernels, choose the configuration offline
(Section~\ref{sec:related}). vLLM's tuner times every
configuration at 18 batch sizes, from 1 to 4,096 tokens, on random router scores that make every
expert equally likely, and keeps the fastest for each. At run time, \emph{vLLM's tuning} uses the
entry nearest the current batch size, on every GPU. DeepGEMM chooses its tiles with a heuristic over the matrix
shape, and FlashInfer profiles its kernels over a range of token counts~\citep{deepgemm,flashinfermoe}.
Each is a function of the batch size alone, shared by every GPU.

\label{sec:problem}
Let $\mathcal{C}$ be the configurations a kernel supports, and $t_c(\mathbf{n}_g)$ the kernel's time
under configuration $c$ on load $\mathbf{n}_g$. The layer waits for the slowest GPU, and each GPU's
configuration affects only its own time, so the layer is fastest when every GPU runs the best
configuration for its own load:
\begin{equation}
t_{\mathrm{layer}} = \max_{g}\, t_{c_g}(\mathbf{n}_g),
\qquad
c_g^{\star} = \operatorname*{arg\,min}_{c \in \mathcal{C}} \; t_c(\mathbf{n}_g).
\label{eq:best}
\end{equation}
Timing every configuration on each GPU's actual load and keeping the fastest finds $c_g^{\star}$;
we call this \emph{exhaustive search}, the oracle against which every choice in this paper is
measured. It is a reference, not a method: it runs every candidate on every batch.
The batch size does not determine $\mathbf{n}_g$: many loads share one $B$, and under EP each GPU
receives a different one. A function of $B$ alone therefore cannot follow a change in $c_g^{\star}$
that routing causes. Not every change in the load moves $c_g^{\star}$, and different configurations
can take almost the same time, so how much this costs is an empirical question. On batches of 128 tokens or more,
giving each GPU its own best configuration is up to 1.45\x faster than vLLM's tuning, and the gap
is largest when each GPU holds few experts (Section~\ref{sec:eval}).

Solving Equation~\ref{eq:best} at run time is constrained in three ways. First, $\mathbf{n}_g$ is
known only after routing, and only on GPU $g$. Second, the choice is made on every forward pass, so
it must take far less time than the kernel, and $t_c(\mathbf{n}_g)$ cannot be measured; it must be
predicted from $\mathbf{n}_g$ alone. Third, serving systems capture each forward pass in a CUDA graph, which records every kernel and
its launch parameters once and replays them, so that the CPU does not launch kernels one by
one~\citep{vllm}. A configuration chosen after routing therefore cannot be a different kernel; the
captured kernel itself must be able to run any $c \in \mathcal{C}$. 

\section{RaMP}
\label{sec:ramp}

\sys solves Equation~\ref{eq:best} on every GPU, after routing, inside the captured forward pass. It replaces measurement with a prediction that only has to rank the
configurations (Section~\ref{sec:costmodel}), makes the choice in the kernel that already groups
tokens by expert (Section~\ref{sec:dispatch}), and runs the choice in one kernel that holds every
candidate (Section~\ref{sec:poly}). Each part answers one constraint of Section~\ref{sec:problem}:
the time must be predicted, the load exists only on the GPU, and the captured forward pass fixes the
kernel.

\subsection{The Cost Model}
\label{sec:costmodel}

Choosing $c_g^{\star}$ exactly requires $t_c(\mathbf{n}_g)$ for every configuration, and only
running the kernel gives it. The obvious substitutes fall short. Timing the candidates on the live
load would cost more than the kernel itself. A per-GPU table, indexed by each GPU's token count, is cheap, but
it treats two loads with the same count as the same work, which Section~\ref{sec:routingbg} showed
they are not. It is also tied to one layout: at another EP degree, the same count spreads over a
different number of experts.

Choosing does not need the time, only the order of the times, and that order is set by a few counts
that the load determines and the configuration fixes. Consider how the kernel spends its time. All
tiles of a configuration have the same shape, and a padded tile does as much work as a full one, so
the work is set by the tiles the kernel launches rather than by the tokens it receives. While the GPU
is full, tiles run $S_c$ at a time, so the time follows the waves $W_c$, and a nearly empty last wave
costs as much as a full one. Each tile also moves a slice of its expert's weights and a block of
token rows, traffic that grows with the tiles $T_c$. That traffic has a floor: every expert that
receives a token must have its weights read at least once, however its tokens are tiled, and this
floor dominates small batches, where the kernel waits on memory. What remains is the fixed cost of a
launch. \sys therefore predicts each configuration's time as
\begin{equation}
\hat{t}_c(\mathbf{n}_g) = \theta_{c,0} + \theta_{c,1}\, W_c(\mathbf{n}_g) + \theta_{c,2}\, T_c(\mathbf{n}_g)
+ \theta_{c,3}\, A(\mathbf{n}_g),
\label{eq:cost}
\end{equation}
where $T_c$ and $W_c$ are the tiles and waves of Equation~\ref{eq:tiles}, and
$A(\mathbf{n}_g) = |\{e \in \mathcal{E}_g : n_e > 0\}|$ is the number of active experts. $q_c$
includes any split of the reduction dimension (split-K), so $T_c$ counts every thread block the
configuration launches.

The counts see exactly what the batch size hides. On the two GPUs of Section~\ref{sec:routingbg},
which both receive 128 tokens, moving from balanced to unbalanced routing raises the row-tiles the 8-row
configuration launches from 16 to 24 and lowers the 32-row configuration's from 16 to 12. The
two move in opposite directions, and a token count sees neither change. The batch size itself does
not appear in Equation~\ref{eq:cost}; it reaches the model only through $\mathbf{n}_g$, so one set
of coefficients serves every batch size and every GPU. Each configuration has its own coefficients,
because its cost per wave, per tile, and per expert depends on its tile shape and pipeline.
Second-order effects that the counts omit, such as the order in which tiles launch, are left to the
fit; excluding the configurations whose launch order the model misjudges changes the geometric
mean by under 0.1\%.

We fit $\theta_c$ offline, from loads $\mathbf{n}^{(i)}$ recorded in serving and the measured time
of $c$ on each, by non-negative least squares~\citep{nnls} on the relative error
$\sum_i \big(\hat{t}_c(\mathbf{n}^{(i)})/t_c(\mathbf{n}^{(i)}) - 1\big)^2$. Both choices follow from
how the model is used. The choice compares configurations on one load, and an error of 2\,$\mu$s
decides it on a 20\,$\mu$s kernel but not on a 500\,$\mu$s one, so error is measured relative to
each load's time. Non-negativity keeps every term a cost, so each prediction grows with every $n_e$, a useful bias since the true time almost always does; an
unconstrained fit gives the correlated wave and tile terms large coefficients of opposite sign that
cancel only on the training loads.

Ranking is also robust to the model's error. An error shared by every configuration on a load, such
as the GPU's clock, the state of its caches, or overhead outside the counts, shifts every prediction
together and leaves the argmin unchanged. On loads from domains it was not fit on, the model orders
the configurations with a rank correlation of 0.97--0.99 with the measured order, while the predicted time of the
configuration it picks is off by a median of 1.4--6.5\% (Section~\ref{sec:eval}; Figure~\ref{fig:costmodel} in the
appendix).

\sys's \emph{menu} $\mathcal{M} \subseteq \mathcal{C}$ holds the top six configurations, chosen greedily
offline to minimize the recorded workload's time, all from one occupancy class
(Section~\ref{sec:poly}).

\subsection{Choosing on the GPU}
\label{sec:dispatch}

The selection kernel takes the router's chosen experts and produces two outputs: the index of the
chosen configuration $\hat{c}_g = \arg\min_{c \in \mathcal{M}} \hat{t}_c(\mathbf{n}_g)$, written to
GPU memory, and the tokens grouped by expert and padded
into that configuration's tiles (Algorithm~\ref{alg:dispatch}). Serving systems already run a
kernel before every MoE kernel that produces the second output for a fixed tile size; \sys
replaces it, so the choice needs no extra launch. One thread block counts this GPU's tokens per
local expert with shared-memory atomics, ignoring tokens routed to other GPUs, and one warp per menu
entry evaluates Equation~\ref{eq:cost}.

For 8 to 4,096 token--expert pairs, the selection kernel takes 3.8--6.5\,$\mu$s and is faster
than the grouping kernel it replaces at 7 of the 8 sizes we measured, so the choice adds under
1\,$\mu$s at most. Its launch shape is fixed and it never waits on the CPU, so it is captured in the
CUDA graph like any other kernel.

\begin{algorithm}[t]
\caption{Selection kernel, one thread block per GPU}
\label{alg:dispatch}
\begin{algorithmic}
\STATE {\bfseries Input:} top-$k$ expert ids, local expert map, menu $\mathcal{M}$, coefficients $\theta$
\STATE $n_e \leftarrow 0$ for each local expert $e$ \hfill // shared memory
\FOR{each token--expert pair $i$ (in parallel)}
  \STATE $e \leftarrow$ local id of the expert of pair $i$; {\bfseries if} $e$ is local {\bfseries then} atomically $n_e \leftarrow n_e + 1$
\ENDFOR
\FOR{each $c \in \mathcal{M}$ (one warp each)}
  \STATE $T_c, A \leftarrow$ warp sums of $q_c\lceil n_e / b_c\rceil$ and $[n_e > 0]$; $W_c \leftarrow \lceil T_c / S_c \rceil$
  \STATE $\hat{t}_c \leftarrow \theta_{c,0} + \theta_{c,1} W_c + \theta_{c,2} T_c + \theta_{c,3} A$
\ENDFOR
\STATE $\hat{c}_g \leftarrow \arg\min_c \hat{t}_c$; write $\hat{c}_g$ to GPU memory
\STATE offsets $\leftarrow$ prefix sum of $b_{\hat{c}_g} \lceil n_e / b_{\hat{c}_g} \rceil$
\STATE scatter token ids into padded tiles of $b_{\hat{c}_g}$ rows
\end{algorithmic}
\end{algorithm}

\subsection{The Kernel}
\label{sec:poly}

The captured forward pass fixes which kernel runs after the selection kernel. One way to branch
anyway is a conditional node, which picks a kernel from a value in GPU memory. Each costs about
8.4\,$\mu$s per layer, or about 0.4\,ms per decode forward pass across Qwen3-30B's 48 MoE layers,
more than the choice could save on a small batch. But the graph fixes only which kernel runs and its
launch shape, not what the kernel does.

\sys therefore compiles every menu entry into one \emph{polymorphic} kernel. Each thread block reads
$\hat{c}_g$ from GPU memory and runs that entry's code. Holding many entries in one kernel has two
costs, and both are small. The grid is sized for the largest tile count any entry can produce, and
blocks beyond the chosen entry's count exit at once, which costs 0.2\%. All entries share one
shared-memory buffer, so the menu is drawn from one occupancy class; within it, each entry runs at
1.00\x its standalone time on a menu of three and at a median of 1.01\x on a menu of six. Six
entries are enough: they capture nearly all of the choice's benefit, within 1--2\% of a menu of
twelve at the MoE layer, while keeping every entry's shared memory and registers close to its own.

\label{sec:ourkernel}
A choice is only as good as what it can choose from. The kernel is written in CuTe
DSL~\citep{cutlass} and computes both expert matrix multiplications and the gated activation in one
launch, in FP8 with $128 \times 128$ block scaling as used by DeepSeek-V3~\citep{deepseekv3}. Its
configuration space covers 2 to 64 rows per tile, tile width, pipeline depth, persistent and grouped
launch orders, and split-K for large reduction dimensions: 98--337 valid configurations per model
and layout.
Appendix~\ref{app:kernel} describes its construction, and we will release the kernel at a later date.

\section{Evaluation}
\label{sec:eval}

This section evaluates \sys on real MoE serving traffic, first end to end in vLLM and then down to
the configuration each GPU chooses. On serving traffic its cost model was never fit on, the configurations it chooses from the full configuration space are up to 1.45\x faster than vLLM's tuning on batches of 128 tokens or more (up to 1.27\x over all batches), and within 3.5\% of exhaustive search on every EP layout. Its kernel and choice together make the MoE layer up to 1.84\x faster than vLLM's Triton MoE on batches of 128 tokens or more, and in vLLM, under TP and EP, it
lowers time per output token at every load we test, by up to 1.11\x over the fastest stock backend.

\textbf{Testbed.} All experiments run on one node with 8 NVIDIA H200 GPUs, connected all-to-all by
NVLink~4, in FP8 with $128 \times 128$ block scaling. End-to-end serving uses CUDA graphs; the Measurement
paragraph gives each microbenchmark's launch mode. The software is vLLM~0.32.0,
PyTorch~2.13, CUDA~13.0, CuTe DSL~4.8, Triton~3.7.1, FlashInfer~0.7.0, and DeepGEMM~2.8.2.
\label{sec:setup}

\textbf{Models and traffic.} Table~\ref{tab:models} lists the eight
models~\citep{olmoe,qwen15moe,qwen2,qwen3moe,qwen35,phi3,mixtral,mixtral8x22b,deepseekv3}; Qwen1.5-MoE enters the tile-size
measurements of Section~\ref{sec:kernel}. We recorded their
routing while vLLM served prompts from four domains: chat~\citep{sharegpt}, math~\citep{gsm8k},
code~\citep{humaneval}, and knowledge questions~\citep{mmlu}, at 16, 64, and 256 concurrent
requests. For seven of the models, we also recorded open-loop traffic, with ShareGPT requests
arriving at random at a fixed rate. Routing does not depend on the layout, so the same recordings
drive every layout; under EP, experts are assigned to GPUs in contiguous blocks, as in vLLM.

\textbf{Measurement.} To compare configurations, we replay recorded batches and time every
configuration on every GPU's actual load, with the cache flushed before each run. The MoE layer's
latency is that of its slowest GPU. EP16 and EP32 replay each GPU's load on the 8 GPUs, without the
exchange of tokens between GPUs, which costs the same under every policy. Table~\ref{tab:openloop} times the expert kernel with direct
launches; Table~\ref{tab:heldout} and Figure~\ref{fig:ladder} replay CUDA graphs, and
Figure~\ref{fig:ladder} also includes \sys's activation quantization and selection kernel. Repeated
measurements differ by 0.2--1.2\%.

\textbf{Baselines.} End to end, we compare with vLLM's three FP8 MoE backends: its Triton
kernel~\citep{triton}, FlashInfer's CUTLASS kernel~\citep{flashinfer,flashinfermoe}, and DeepGEMM~\citep{deepgemm};
where vLLM ships no tuned table for a model's shape, its Triton kernel runs its default configuration.
vLLM's own CUTLASS FP8 MoE backend does not support these configurations, so it is not compared.
To isolate the choice from the kernel, we also run \sys's kernel under four other policies.
vLLM's tuning (Section~\ref{sec:today}) is applied to \sys's kernel and
configuration space. The \emph{trace-tuned table} is the same lookup tuned on real traffic. The \emph{per-GPU table} is indexed by each GPU's own token count. And
exhaustive search (Section~\ref{sec:problem}) times every configuration on each GPU and keeps the
fastest. Policies that learn from traffic never see the traffic
they are scored on.

\begin{table}[t]
\caption{\textbf{Models.} Total and active parameters, experts per layer $E$, experts per token
$k$, and expert intermediate size $h$.}
\label{tab:models}
\vskip 0.08in
\centering
\footnotesize
\setlength{\tabcolsep}{4.5pt}
\begin{tabular}{lcrrr}
\toprule
Model & Params & $E$ & $k$ & $h$ \\
\midrule
OLMoE         & 7B / 1B    & 64  & 8 & 1,024  \\
Qwen1.5-MoE   & 14B / 2.7B & 60  & 4 & 1,408  \\
Qwen2-57B & 57B / 14B  & 64  & 8 & 2,560  \\
Qwen3-30B & 30B / 3B   & 128 & 8 & 768    \\
Qwen3.5-35B & 35B / 3B   & 256 & 8 & 512    \\
Phi-3.5-MoE   & 42B / 6.6B & 16  & 2 & 6,400  \\
Mixtral-8x22B & 141B / 39B & 8   & 2 & 16,384 \\
DeepSeek-V3   & 671B / 37B & 256 & 8 & 2,048  \\
\bottomrule
\end{tabular}
\vskip -0.1in
\end{table}

\subsection{End-to-End Serving}
\label{sec:e2e}

We served DeepSeek-V3 and Qwen3-30B on EP8 and OLMoE on EP4, with attention split over the same
GPUs, and DeepSeek-V3 also on TP8. Requests come from ShareGPT, at a fixed rate or a fixed number of
concurrent requests. For Qwen3-30B, the menu and cost model were fit without the chat domain, so the
served traffic is held out.

\begin{table*}[t]
\caption{\textbf{End-to-end serving in vLLM.} Mean time per output token (ms, median of three runs;
lower is better) and \sys's speedup over each stock MoE backend. Load: request rate or concurrent
requests.}
\label{tab:e2e}
\vskip 0.08in
\centering
\footnotesize
\setlength{\tabcolsep}{7pt}
\begin{tabular}{l rrrr @{\hspace{1.6em}} rrr}
\toprule
& \multicolumn{4}{c}{Time per output token (ms)} & \multicolumn{3}{c}{\sys's speedup over} \\
\cmidrule(lr){2-5}\cmidrule(l){6-8}
Load & Triton & FlashInfer & DeepGEMM & \sys & Triton & FlashInfer & DeepGEMM \\
\midrule
\multicolumn{8}{l}{\textit{DeepSeek-V3, EP8}} \\
4 req/s & 20.15 & 19.98 & 23.40 & \textbf{18.91} & \textbf{1.07\x} & \textbf{1.06\x} & \textbf{1.24\x} \\
8 req/s & 28.60 & 28.93 & 32.99 & \textbf{26.63} & \textbf{1.07\x} & \textbf{1.09\x} & \textbf{1.24\x} \\
16 req/s & 29.76 & 31.89 & 32.06 & \textbf{27.34} & \textbf{1.09\x} & \textbf{1.17\x} & \textbf{1.17\x} \\
64 conc. & 30.08 & 30.07 & 31.32 & \textbf{27.73} & \textbf{1.08\x} & \textbf{1.08\x} & \textbf{1.13\x} \\
256 conc. & 38.33 & 41.71 & 40.21 & \textbf{37.92} & \textbf{1.01\x} & \textbf{1.10\x} & \textbf{1.06\x} \\
\addlinespace[2pt]
\multicolumn{8}{l}{\textit{DeepSeek-V3, TP8}} \\
64 conc. & 27.79 & 40.50 & 38.71 & \textbf{25.83} & \textbf{1.08\x} & \textbf{1.57\x} & \textbf{1.50\x} \\
256 conc. & 37.54 & 57.65 & 49.86 & \textbf{35.75} & \textbf{1.05\x} & \textbf{1.61\x} & \textbf{1.39\x} \\
\addlinespace[2pt]
\multicolumn{8}{l}{\textit{Qwen3-30B, EP8, held-out domain}} \\
8 req/s & 5.23 & 4.87 & 5.63 & \textbf{4.51} & \textbf{1.16\x} & \textbf{1.08\x} & \textbf{1.25\x} \\
16 req/s & 5.23 & 4.89 & 5.64 & \textbf{4.50} & \textbf{1.16\x} & \textbf{1.09\x} & \textbf{1.25\x} \\
32 req/s & 5.55 & 5.35 & 6.03 & \textbf{4.82} & \textbf{1.15\x} & \textbf{1.11\x} & \textbf{1.25\x} \\
64 conc. & 5.91 & 5.79 & 6.34 & \textbf{5.28} & \textbf{1.12\x} & \textbf{1.10\x} & \textbf{1.20\x} \\
128 conc. & 7.10 & 7.05 & 7.74 & \textbf{6.46} & \textbf{1.10\x} & \textbf{1.09\x} & \textbf{1.20\x} \\
\addlinespace[2pt]
\multicolumn{8}{l}{\textit{OLMoE, EP4}} \\
32 req/s & 2.77 & 2.56 & 2.98 & \textbf{2.39} & \textbf{1.16\x} & \textbf{1.07\x} & \textbf{1.25\x} \\
64 conc. & 3.05 & 2.90 & 3.16 & \textbf{2.78} & \textbf{1.10\x} & \textbf{1.04\x} & \textbf{1.14\x} \\
192 conc. & 3.98 & 3.93 & 3.99 & \textbf{3.90} & \textbf{1.02\x} & \textbf{1.01\x} & \textbf{1.02\x} \\
\bottomrule
\end{tabular}
\vskip -0.1in
\end{table*}

Table~\ref{tab:e2e} shows that \sys is the fastest MoE backend at every load, against each stock
backend. On Qwen3-30B, serving a domain its cost model never saw, it is 1.10--1.16\x faster than
Triton, 1.08--1.11\x than FlashInfer, and 1.20--1.25\x than DeepGEMM. On DeepSeek-V3 it is
1.01--1.24\x faster than every stock backend on EP8 and 1.05--1.61\x on TP8, and on OLMoE
1.01--1.25\x.
Model quality is unchanged: perplexity on 5,000--5,700 tokens of Python standard-library documentation is
1.124--1.126 on DeepSeek-V3 and 5.50--5.52 on OLMoE across all backends, and, on Qwen3-30B serving the held-out domain, 3.02 with \sys
against 3.05 with FlashInfer.

The gain holds even where the MoE layer is a small part of each step. On Qwen3-30B, the experts take
only 15\% of GPU time and communication most of the rest, yet \sys is 1.08--1.11\x faster per output
token. Part of that communication is GPUs waiting for the slowest GPU's MoE layer, and it falls from
62\% of the step with FlashInfer to 57\% with \sys.

On one node, the room for any choice is bounded: over all batches of the recorded traffic, exhaustive search is 1.04--1.07\x faster than vLLM's tuning on \sys's kernel, which caps what a choice can
add to time per output token at 1--3\%. \sys's menu is itself chosen from recorded routing, so a
batch-size table on that menu already carries part of the choice, and \sys's live choice is between 2\% slower and 5\% faster at every point (Table~\ref{tab:e2e-policies}). The gain from the choice grows with the number of GPUs
and with the share of large batches, as Section~\ref{sec:eval-source} shows: on open-loop traffic at
EP32, \sys is 1.18--1.22\x faster than vLLM's tuning on the same kernel.

\subsection{Where the Speedup Comes From}
\label{sec:eval-source}

The end-to-end gain has two sources, \sys's kernel and the configuration it chooses for each GPU.
We separate them at the MoE layer, then test the choice on its own.

\begin{figure*}[t]
\centering
\includegraphics[width=\textwidth]{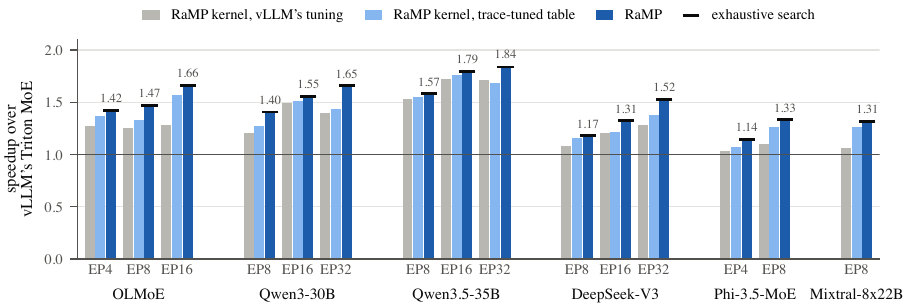}
\vskip -0.1in
\caption{\textbf{MoE layer speedup over vLLM's Triton MoE under EP}, on batches of 128 tokens or more
from held-out domains. Bars: \sys's kernel under vLLM's tuning, under the trace-tuned table, and
with \sys's choice; ticks: exhaustive search. Latency is the slowest GPU's, including \sys's
activation quantization and selection kernel, so the ratios are lower than the expert-kernel times of
Tables~\ref{tab:openloop} and~\ref{tab:heldout}. EP32 bars use a smaller sample of batches than the starred rows of Table~\ref{tab:heldout}.}
\label{fig:ladder}
\end{figure*}

\textbf{Kernel and choice.} Figure~\ref{fig:ladder} builds up \sys's MoE layer latency under EP from
vLLM's Triton MoE, one change at a time, on batches of 128 tokens or more. \sys's kernel under
vLLM's tuning is 1.03--1.72\x faster than Triton. Choosing its configuration from each GPU's load
adds another 1.03--1.29\x, for a total of 1.14--1.84\x, including the selection kernel. The choice
is a median of 40\% of the total gain, and up to 79\% on models with few, large experts such as
Mixtral-8x22B and Phi-3.5-MoE.

\textbf{Choosing on serving traffic.} Table~\ref{tab:openloop} tests the choice on open-loop serving
traffic. Every policy is fit on the recorded traffic and scored on batches sampled from open-loop
serving, which it never saw. On batches of 128 tokens or more, \sys's configurations are
1.03--1.45\x faster than vLLM's tuning; over all batches, 1.01--1.27\x, within 0.7--3.5\% of
exhaustive search on all 14 EP layouts and within 2\% on 8 of them. Against the trace-tuned table,
\sys is 0.99--1.08\x over all batches. Table~\ref{tab:openloop} chooses from \sys's full
configuration space. Serving uses the top six configurations, and on them \sys is 1.06\x faster than
a table tuned on the same six (geometric mean over 21 EP layouts; up to 1.27\x) and on average
within 0.5\% of the best of the six (Table~\ref{tab:menu}). Against the full space, the served
kernel is 1.12\x faster than vLLM's tuning (geometric mean; up to 1.68\x) and on average 7\% from exhaustive search over all configurations; what remains is the cost of serving six configurations,
not of choosing among them.

\begin{table}[t]
\caption{\textbf{Choosing on open-loop serving traffic.} Speedup of \sys over each policy, all on
\sys's kernel and fit on the recorded traffic, over all batches and on batches of 128 tokens or
more ($\geq$128); the trace-tuned and per-GPU columns are over all batches. Times are those of the
expert kernel. Gap: \sys's time over exhaustive search, over all batches (\%).}
\label{tab:openloop}
\vskip 0.08in
\centering
\footnotesize
\setlength{\tabcolsep}{2.6pt}
\begin{tabular}{llrrrrr}
\toprule
& & \multicolumn{2}{c}{vLLM's tuning} & & & \\
\cmidrule(lr){3-4}
Model & Layout & All & $\geq$128 & \begin{tabular}[b]{@{}c@{}}Trace-\\tuned\end{tabular} & \begin{tabular}[b]{@{}c@{}}Per-GPU\\table\end{tabular} & Gap \\
\midrule
Qwen3-30B & EP8 & 1.10\x & 1.24\x & 1.04\x & 1.03\x & 3.5 \\
 & EP16 & 1.04\x & 1.06\x & 1.02\x & 1.01\x & 2.2 \\
 & EP32 & 1.18\x & 1.24\x & 1.06\x & 1.00\x & 2.0 \\
Qwen3.5-35B & EP8 & 1.05\x & 1.07\x & 1.02\x & 1.02\x & 2.6 \\
 & EP32 & 1.18\x & 1.24\x & 1.07\x & 1.01\x & 2.1 \\
DeepSeek-V3 & EP8 & 1.01\x & 1.03\x & 0.99\x & 0.99\x & 2.9 \\
 & EP32 & 1.22\x & 1.16\x & 1.03\x & 1.01\x & 1.0 \\
OLMoE & EP8 & 1.05\x & 1.16\x & 1.03\x & 1.02\x & 0.9 \\
 & EP16 & 1.14\x & 1.45\x & 1.00\x & 1.00\x & 2.0 \\
Phi-3.5-MoE & EP8 & 1.27\x & 1.27\x & 1.06\x & 1.03\x & 1.0 \\
 & EP16 & 1.20\x & 1.34\x & 1.08\x & 1.01\x & 2.9 \\
Qwen2-57B & EP8 & 1.04\x & 1.04\x & 1.01\x & 1.01\x & 1.2 \\
 & EP16 & 1.17\x & 1.29\x & 1.04\x & 1.02\x & 1.4 \\
Mixtral-8x22B & EP8 & 1.14\x & 1.14\x & 1.03\x & 1.00\x & 0.7 \\
\bottomrule
\end{tabular}
\vskip -0.1in
\end{table}

What pays is choosing from each GPU's own load, which no policy keyed on the batch size can do. Even the simplest
such policy, a per-GPU table, needs \sys's mechanism, since the count exists only on the GPU after
routing; at the EP degree it was tuned for, it comes within 3\% of \sys. The cost model adds two
things a table cannot. How the tokens spread over a GPU's experts adds up to 1.07\x on batches of
128 tokens or more at EP16 and EP32. And the cost model carries over to new layouts: a per-GPU table
tuned at a model's other EP degrees is 1.09\x from exhaustive search, against 1.04\x at its own
degree, while the cost model, with the layer's dimensions added to its terms and fit on seven
models, is within 1.9\% of exhaustive search on the eighth (geometric mean).

\textbf{Why the cost model works.} It ranks configurations almost exactly, with a rank correlation of
0.97--0.99 with the measured order, which is all the choice uses. Its terms matter: a formula built
from the tile count alone (the tiles, the tiles per SM, and their logarithm, fit by ordinary least
squares) is 6.7\% from exhaustive search on batches of 128 tokens or more, against 1.3\% for
Equation~\ref{eq:cost} (geometric mean over 18 EP layouts). Nor can a table keyed on the same inputs
replace it: over the 21 EP layouts of Appendix~\ref{app:tables}, tables keyed on a GPU's token
count and a measure of its balance, such as its padding or active experts, are 3.1--4.8\% from
exhaustive search on these batches, against 1.4\% for the cost model,
because a few buckets lose how each expert's tokens fill tiles and waves. The model's rare mispicks fall on GPUs that seldom set the layer's latency;
removing them changes the result by under 0.1\%.

\textbf{When the choice matters.} When every GPU holds every expert, each GPU computes the whole batch's routing,
and at a given batch size the best configuration changes little. This is the case offline
tuning was built for, and \sys's choice matches the trace-tuned table to within 2\%, while its MoE
layer is 1.09--1.17\x faster than vLLM's Triton MoE (Table~\ref{tab:tp}). Balanced expert placement
does not remove the imbalance under EP, because the most loaded experts change from batch to batch.
It lowers the imbalance by at most 11 percentage points, and \sys's gain is unchanged, within its 95\%
interval, on all six layouts we test (Table~\ref{tab:balanced}). Even re-balancing every 128
forward passes leaves the most loaded GPU 11--56\% above the mean on EP8 (Table~\ref{tab:placement}).

The choice pays most as each GPU holds fewer experts. On Qwen3-30B, the most loaded GPU holds 52\% more
than the mean at EP8 and 188\% more at EP32, and \sys's gain over the trace-tuned table on held-out
batches of 128 tokens or more grows from 1.14\x to 1.32\x; on DeepSeek-V3, the excess grows from
45\% to 132\% and the gain from 1.00\x to 1.15\x (Tables~\ref{tab:imbalance} and~\ref{tab:heldout}). It also grows with
the share of large batches, which open-loop traffic carries far more of than the recorded traffic: on
Qwen3-30B, 14--52\% of forward passes carry 128 tokens or more under open-loop arrivals, against
4\%. Weighted by the open-loop mix at 32 requests per second, \sys's gain over vLLM's tuning on the
MoE layer is 1.16--1.23\x on four of six EP layouts, close to that of exhaustive search.

\section{Related Work}
\label{sec:related}

\textbf{MoE kernels.} MoE megakernels and grouped kernels include vLLM's Triton MoE~\citep{vllm,triton},
MegaBlocks~\citep{megablocks}, ScatterMoE~\citep{scattermoe}, DeepGEMM~\citep{deepgemm},
FlashInfer's CUTLASS path~\citep{flashinfermoe,cutlass}, Alpha-MoE~\citep{alphamoe}, and
SonicMoE~\citep{sonicmoe}. They differ in how they gather tokens, fuse the activation, and overlap
memory with compute; MegaBlocks, for example, casts the layer as a block-sparse product with a
fixed $128 \times 128$ block, chosen by benchmarking dense matrix multiplications before any routing
exists. These works make one kernel faster and choose its configuration
statically or from the batch size. \sys instead chooses which configuration to run for each GPU on
every forward pass, and the same choice applies to any kernel that exposes several configurations.

\textbf{Load balance and placement.} Training adds capacity limits and auxiliary losses that push
the router toward balance~\citep{gshard,switch}; FasterMoE replicates popular experts during
training~\citep{fastermoe}, and Tutel switches the parallelism strategy at every iteration to follow
the changing load~\citep{tutel}. At serving time, DeepSeek-V3 replicates popular
experts~\citep{deepseekv3,eplb}, and \citet{huang2023moedeploy} buffer and rebalance experts. These methods reduce the imbalance between GPUs by deciding where experts
and tokens go. \sys handles what they leave: the imbalance that remains from batch to batch, which
balanced placement lowers by at most 11 percentage points, and the imbalance among each GPU's own experts, which no
placement sees. It runs on whatever placement they produce.

\textbf{Autotuning and run-time kernel selection.} Halide, TVM, Ansor, and AutoTVM search
configuration spaces offline, guided by measurements or a learned cost
model~\citep{halide,tvmosdi,tvm,autotvm}; like
\sys's, AutoTVM's model needs only to rank candidates. Triton autotunes kernels for each input
shape~\citep{triton}, and DeepGEMM chooses tiles from the matrix shape~\citep{deepgemm}.
DietCode~\citep{dietcode} is the closest: it compiles micro-kernels for dynamic shapes, uses a
cost model offline to choose among them for each shape, and dispatches at run time with a decision
tree trained on those choices. In all of these, the input to the choice is a
shape that the CPU knows before the kernel launches. For an MoE layer, the deciding input is a
vector, each GPU's per-expert load, and it exists only on that GPU, after routing, in the middle of
the captured forward pass. \sys therefore evaluates a closed-form ranking model on the GPU and
switches configurations inside one polymorphic kernel, with no CPU involved.

\textbf{Serving and batching.} Continuous batching~\citep{orca}, paged attention~\citep{vllm},
chunked prefill~\citep{sarathi}, and structured programs~\citep{sglang} decide how many tokens each
forward pass carries. They treat the batch size as the kernel's problem size; for an MoE layer, the
router changes the problem again before the kernel runs.

\section{Limitations}
\label{sec:limits}

\sys is evaluated on one node of eight H200 GPUs. EP16 and EP32 replay each GPU's recorded load
without the exchange of tokens between GPUs; serving them across nodes, where each GPU holds the
fewest experts and the choice pays most, is the natural next step. On one node, \sys's end-to-end gain comes mostly from its kernel; the
choice pays most at higher EP degrees, which we evaluate by replay. We evaluate FP8 with $128 \times 128$ block scaling in
vLLM; other precisions, GPU generations, and serving systems are left to future work.

The cost model counts the tiles, waves, and active experts a configuration produces, but a kernel
can apply further optimizations whose cost these counts do not capture: the reduction of split-K's
partial results, tile orders chosen for L2 reuse, separate tile shapes for the up and down
projections, and two-CTA matrix multiplications on newer GPUs. How the cost model should account
for these optimizations, and how they interact with routing, is a direction for future research.

\section{Conclusion}
\label{sec:conclusion}

Serving systems choose an MoE kernel's configuration from the batch size, but the router decides how
the batch's tokens split across experts and GPUs, and the best configuration follows that split.
\sys makes the choice where the split is known: on each GPU, after routing, on every forward pass. A
cost model ranks the configurations from each GPU's load, and a polymorphic kernel runs whichever
one it picks. On serving traffic the model was never fit on, the configurations it chooses from the full configuration space are up to 1.45\x
faster than vLLM's tuning on batches of 128 tokens or more and within 3.5\% of exhaustive search over all batches,
and in vLLM \sys lowers time per output token by up to 1.11\x over the fastest stock backend under
both TP and EP. As expert-parallel serving spreads models over more GPUs, each GPU holds fewer
experts, its share of the batch grows less balanced, and choosing from routing pays more.

\bibliography{references}
\bibliographystyle{mlsys2025}

\section*{AI Usage Disclosure}
\label{sec:ai}

We used generative AI tools to help check the claims, refine wording, organize the
experiments, and draft the manuscript. We did not use these tools for translation, transcription, or
qualitative data analysis. The authors ran all reported experiments and model evaluations, tested the
code, verified numerical claims against preserved artifacts, manually reviewed the cited literature,
and reviewed all AI-assisted content. We take responsibility for the final text, claims, code, figures,
and experimental results.

\clearpage
\appendix

\onecolumn
\section{Additional Results}
\label{sec:appendix}

\subsection{The Cost Model on Held-Out Loads}
\label{app:costmodel}

\begin{figure}[h!]
\centering
\includegraphics[width=\textwidth]{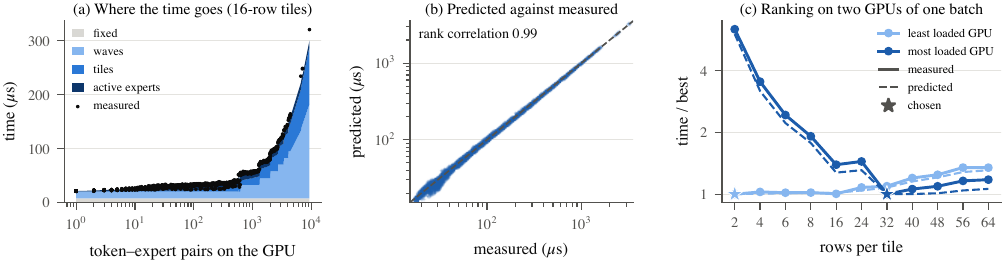}
\vskip -0.1in
\caption{\textbf{The cost model on held-out loads}, Qwen3-30B on EP8. (a) Each term's share of the
predicted time for one configuration, and the measured time. (b) Predicted against measured time
for every configuration. (c) Measured and predicted time of each tile size on the least and most
loaded GPU of one batch, relative to the best; stars mark the configuration the model picks.}
\label{fig:costmodel}
\end{figure}

Figure~\ref{fig:costmodel} shows the cost model of Section~\ref{sec:costmodel} on loads from a
domain it was not fit on, for Qwen3-30B on EP8. Its terms account for the measured time across four
orders of magnitude of load: the waves carry most of the time and set its steps, the tiles add the
traffic that grows with the load, and the fixed cost and the active experts set the floor at small
loads (a). Over every configuration, its predictions order the measured times with a rank
correlation of 0.99 (b). On two GPUs of one batch, it picks the fastest tile size on both: 2-row tiles for the least loaded GPU
and 32-row tiles for the most loaded one (c).

\subsection{Tables Given the Cost Model's Inputs}
\label{app:tables}

Table~\ref{tab:tables} replaces the cost model with tables that see the same per-GPU information:
the GPU's token count crossed with a measure of how the tokens spread over its experts. The best
table, keyed on the token count and the padding, is 3.1\% from exhaustive search on batches of 128
tokens or more and 5.8\% on its worst layout, against 1.4\% and 2.9\% for the cost model. A
configuration's time depends on how each expert's tokens fill tiles and waves, which a few buckets
lose, and at large batch sizes each bucket holds few recorded batches.

\begin{table}[h!]
\caption{\textbf{Tables given the cost model's inputs.} Gap of each per-GPU policy to exhaustive
search on batches of 128 tokens or more (\%; geometric mean over 21 EP layouts, and the worst
layout), held-out domains, direct launches, full configuration space. Over all batches, where most
GPUs receive few tokens and the configuration matters little, every policy is within 2.2\%.}
\label{tab:tables}
\vskip 0.08in
\centering
\footnotesize
\setlength{\tabcolsep}{6pt}
\begin{tabular}{llcc}
\toprule
Policy & Keyed on & Gap & Worst layout \\
\midrule
Per-GPU table & token count & 4.0 & 8.7 \\
2-D table & token count $\times$ padding (8 buckets) & 3.1 & 5.8 \\
2-D tables & token count $\times$ active experts, largest load over mean, & 3.3--4.8 & -- \\
 & \quad or coefficient of variation (2--8 buckets) & & \\
3-D table & token count $\times$ active experts $\times$ padding (quartiles) & 3.4 & 6.4 \\
8 nearest neighbors & token count, active experts, padding, largest load over mean & 3.5 & 8.6 \\
\midrule
Cost model & the GPU's tokens per expert & \textbf{1.4} & \textbf{2.9} \\
\bottomrule
\end{tabular}
\end{table}

\subsection{Choosing on Held-Out Domains}
\label{app:heldout}

Table~\ref{tab:heldout} complements Table~\ref{tab:openloop} with recorded traffic: every policy is
fit on three domains and scored on batches of 128 tokens or more from the fourth. \sys is within
2\% of exhaustive search on every layout, and its gain over the trace-tuned table is largest at
EP32 on every model we evaluate there.

\begin{table}[h!]
\caption{\textbf{Held-out domains.} Expert-kernel speedup under EP on batches of 128 tokens or more
from a domain each policy was not fit on, over vLLM's tuning and over the trace-tuned table. Latency
is the slowest GPU's. Brackets: 95\% bootstrap interval. Every policy runs \sys's kernel and
chooses from its full configuration space. $^*$Measured on a 2--2.5\x larger sample of batches. Differences under 1\% between \sys and
exhaustive search are within timing noise.}
\label{tab:heldout}
\vskip 0.08in
\centering
\footnotesize
\setlength{\tabcolsep}{5pt}
\begin{tabular}{ll@{\hspace{2em}}cccc@{\hspace{2em}}cc}
\toprule
& & \multicolumn{4}{c}{Over vLLM's tuning} & \multicolumn{2}{c}{Over trace-tuned table} \\
\cmidrule(r{2em}){3-6}\cmidrule{7-8}
Model & Layout & Trace-tuned & Per-GPU table & \sys & Exhaustive & \sys [95\% CI] & Exhaustive \\
\midrule
OLMoE       & EP4  & 1.10 & 1.11 & 1.15 & 1.15 & 1.04 [1.02, 1.07] & 1.05 \\
            & EP8  & 1.08 & 1.16 & 1.24 & 1.23 & 1.15 [1.11, 1.18] & 1.14 \\
            & EP16$^*$ & 1.11 & 1.39 & 1.40 & 1.41 & 1.27 [1.20, 1.33] & 1.27 \\
Qwen3-30B   & EP8  & 1.07 & 1.16 & 1.22 & 1.23 & 1.14 [1.11, 1.16] & 1.15 \\
            & EP16$^*$ & 0.93 & 1.04 & 1.06 & 1.07 & 1.15 [1.10, 1.20] & 1.16 \\
            & EP32$^*$ & 1.04 & 1.36 & 1.38 & 1.38 & 1.32 [1.24, 1.39] & 1.32 \\
Qwen3.5-35B & EP8  & 1.02 & 1.01 & 1.04 & 1.05 & 1.01 [1.00, 1.02] & 1.02 \\
            & EP16$^*$ & 1.04 & 1.05 & 1.08 & 1.08 & 1.03 [1.02, 1.06] & 1.04 \\
            & EP32$^*$ & 1.04 & 1.13 & 1.12 & 1.13 & 1.08 [1.05, 1.13] & 1.09 \\
DeepSeek-V3 & EP8$^*$  & 1.08 & 1.09 & 1.08 & 1.10 & 1.00 [0.99, 1.01] & 1.01 \\
            & EP16$^*$ & 1.01 & 1.01 & 1.08 & 1.09 & 1.07 [1.05, 1.09] & 1.08 \\
            & EP32$^*$ & 1.07 & 1.18 & 1.23 & 1.24 & 1.15 [1.12, 1.18] & 1.15 \\
Phi-3.5-MoE & EP4  & 1.05 & 1.08 & 1.12 & 1.13 & 1.07 [1.04, 1.09] & 1.07 \\
            & EP8  & 1.19 & 1.20 & 1.25 & 1.26 & 1.06 [1.04, 1.07] & 1.06 \\
Mixtral-8x22B & EP8 & 1.20 & 1.24 & 1.26 & 1.26 & 1.05 [1.02, 1.07] & 1.05 \\
\bottomrule
\end{tabular}
\end{table}

\subsection{Imbalance and Expert Placement}
\label{app:imbalance}

Table~\ref{tab:imbalance} gives the imbalance behind Section~\ref{sec:eval-source}: the most loaded
GPU's share grows with the number of GPUs on every model. Table~\ref{tab:placement} shows that
balancing the placement of experts does not remove it, because the most loaded experts change from
batch to batch, and Table~\ref{tab:balanced} shows that \sys's gain holds on a balanced placement.

\begin{table}[h!]
\caption{\textbf{Imbalance by layout.} Most loaded GPU's tokens over the mean on batches of 128
tokens or more, contiguous placement (median). Table~\ref{tab:placement} uses batches of 128--511
tokens, so its contiguous column differs slightly.}
\label{tab:imbalance}
\vskip 0.08in
\centering
\footnotesize
\begin{tabular}{lcccc}
\toprule
Model & EP4 & EP8 & EP16 & EP32 \\
\midrule
OLMoE         & 1.22 & 1.45 & 1.89 & -- \\
Qwen3-30B & --   & 1.52 & 1.95 & 2.88 \\
Qwen3.5-35B & --   & 1.42 & 1.90 & 2.69 \\
DeepSeek-V3   & --   & 1.45 & 1.94 & 2.32 \\
Phi-3.5-MoE   & 1.40 & 1.90 & --   & -- \\
Mixtral-8x22B & --   & 1.54 & --   & -- \\
\bottomrule
\end{tabular}
\end{table}

\begin{table}[h!]
\caption{\textbf{Imbalance under expert placement.} Most loaded GPU over the mean on EP8, under four placements, batches
of 128--511 tokens (median). Offline: balanced on the other domains. Every 128: re-balanced from the previous 128
forward passes. Uniform: uniformly random routing, for reference.}
\label{tab:placement}
\vskip 0.08in
\centering
\footnotesize
\setlength{\tabcolsep}{4pt}
\begin{tabular}{lcccc}
\toprule
Model & Uniform & Contiguous & Offline & Every 128 \\
\midrule
Qwen3-30B & 1.10 & 1.49 & 1.46 & 1.21 \\
Qwen3.5-35B & 1.11 & 1.39 & 1.35 & 1.22 \\
DeepSeek-V3 & 1.13 & 1.40 & 1.30 & 1.16 \\
OLMoE       & 1.10 & 1.43 & 1.34 & 1.16 \\
Phi-3.5-MoE & 1.21 & 1.90 & 1.91 & 1.56 \\
Qwen2-57B   & 1.11 & 1.17 & 1.11 & 1.11 \\
\bottomrule
\end{tabular}
\end{table}

\begin{table}[h!]
\caption{\textbf{\sys on balanced placement.} Experts placed per layer to even out each GPU's
expected load on the other domains, batches of 128 tokens or more. Imbalance before and after
balancing, and \sys's speedup over the trace-tuned table under each placement (brackets: 95\%
bootstrap interval). EP32 rows use a smaller sample of batches than Table~\ref{tab:heldout}.}
\label{tab:balanced}
\vskip 0.08in
\centering
\footnotesize
\setlength{\tabcolsep}{5pt}
\begin{tabular}{llcccc}
\toprule
& & \multicolumn{2}{c}{Imbalance} & \multicolumn{2}{c}{\sys over trace-tuned table} \\
\cmidrule(lr){3-4}\cmidrule(l){5-6}
Model & Layout & Contiguous & Balanced & Contiguous & Balanced \\
\midrule
OLMoE       & EP8  & 1.45 & 1.40 & 1.15 & 1.13 [1.10, 1.16] \\
Qwen3-30B   & EP8  & 1.52 & 1.53 & 1.14 & 1.13 [1.10, 1.16] \\
Qwen3.5-35B & EP8  & 1.42 & 1.36 & 1.01 & 1.01 [1.00, 1.02] \\
DeepSeek-V3 & EP8  & 1.45 & 1.35 & 1.00 & 0.99 [0.99, 1.00] \\
Qwen3-30B   & EP32 & 2.88 & 2.77 & 1.24 & 1.24 [1.11, 1.36] \\
DeepSeek-V3 & EP32 & 2.32 & 2.22 & 1.12 & 1.13 [1.10, 1.17] \\
\bottomrule
\end{tabular}
\end{table}

\subsection{Choosing Among the Served Configurations}
\label{app:menu}

Table~\ref{tab:menu} compares \sys with a batch-size table on the six configurations \sys serves.
On these six, \sys matches or beats the table on 20 of the 21 EP layouts, by up to 1.27\x, and is
within 0.5\% of the best of the six on average and within 3\% on every layout. Against the full configuration
space, the served kernel is 1.12\x faster than vLLM's tuning (geometric mean), faster on 18 of the 21 layouts and by up to 1.68\x, and on average 7\% from exhaustive search over all configurations.

\begin{table}[h!]
\caption{\textbf{Choosing among the six served configurations.} Batches of 128 tokens or more from
held-out domains, direct launches. \sys's speedup over a batch-size table tuned on the same six
configurations, and \sys's gap to the best of the six (\%). Against the full space: the served
kernel's speedup over vLLM's tuning on all configurations, and its gap to exhaustive search over all
configurations (\%).}
\label{tab:menu}
\vskip 0.08in
\centering
\footnotesize
\setlength{\tabcolsep}{6pt}
\begin{tabular}{llcccc}
\toprule
& & \multicolumn{2}{c}{Among the six} & \multicolumn{2}{c}{Against the full space} \\
\cmidrule(lr){3-4}\cmidrule(l){5-6}
Model & Layout & \sys over the table & Gap to the best (\%) & \sys over vLLM's tuning & \begin{tabular}[b]{@{}c@{}}Gap to exhaustive\\search (\%)\end{tabular} \\
\midrule
DeepSeek-V3 & EP8 & 1.01\x & 0.1 & 1.07\x & 3.6 \\
 & EP16 & 0.99\x & 2.8 & 1.02\x & 7.9 \\
 & EP32 & 1.12\x & 0.3 & 1.13\x & 8.2 \\
Mixtral-8x22B & EP4 & 1.00\x & 0.0 & 1.03\x & 1.4 \\
 & EP8 & 1.00\x & 0.0 & 1.23\x & 1.9 \\
OLMoE & EP4 & 1.02\x & 0.6 & 1.04\x & 10.4 \\
 & EP8 & 1.16\x & 0.0 & 1.17\x & 5.5 \\
 & EP16 & 1.03\x & 0.1 & 1.35\x & 8.5 \\
 & EP32 & 1.27\x & 0.4 & 1.68\x & 4.9 \\
Phi-3.5-MoE & EP4 & 1.05\x & 0.1 & 1.01\x & 11.5 \\
 & EP8 & 1.05\x & 0.1 & 1.13\x & 11.6 \\
 & EP16 & 1.16\x & 0.2 & 1.16\x & 11.1 \\
Qwen2-57B & EP8 & 1.02\x & 0.2 & 1.00\x & 7.4 \\
 & EP16 & 1.04\x & 0.0 & 1.18\x & 7.3 \\
Qwen3-30B & EP8 & 1.10\x & 0.0 & 1.17\x & 6.1 \\
 & EP16 & 1.02\x & 0.1 & 1.04\x & 3.2 \\
 & EP32 & 1.09\x & 1.5 & 1.19\x & 10.8 \\
Qwen3.5-35B & EP4 & 1.00\x & 0.2 & 0.98\x & 7.7 \\
 & EP8 & 1.05\x & 0.7 & 0.97\x & 9.7 \\
 & EP16 & 1.01\x & 0.5 & 1.01\x & 6.6 \\
 & EP32 & 1.13\x & 0.8 & 1.13\x & 3.4 \\
\bottomrule
\end{tabular}
\end{table}

\subsection{The Choice on a Fixed Menu}
\label{app:policies}

Table~\ref{tab:e2e-policies} serves \sys's kernel and menu end to end with two ways of choosing from
the menu: \sys's live choice and a batch-size table. Because the menu itself is chosen from recorded
routing, both carry part of \sys's choice, and on one node \sys is between 2\% slower and 5\% faster than the table.

\begin{table}[h!]
\caption{\textbf{\sys's choice end to end.} Time per output token (ms, median of three runs) of \sys
and of a vLLM-style batch-size table on \sys's menu, which is itself chosen from recorded routing
(DeepSeek-V3 TP8: the trace-tuned table on that menu).}
\label{tab:e2e-policies}
\vskip 0.08in
\centering
\footnotesize
\setlength{\tabcolsep}{4pt}
\begin{tabular}{llrr}
\toprule
Model & Load & \begin{tabular}[b]{@{}c@{}}Table on\\\sys's menu\end{tabular} & \sys \\
\midrule
DeepSeek-V3, EP8 & 4 req/s   & 19.41 & 18.91 \\
                 & 8 req/s   & 27.95 & 26.63 \\
                 & 16 req/s  & 27.54 & 27.34 \\
                 & 64 conc.  & 27.65 & 27.73 \\
                 & 256 conc. & 38.55 & 37.92 \\
DeepSeek-V3, TP8 & 64 conc.  & 25.83 & 25.83 \\
                 & 256 conc. & 35.76 & 35.75 \\
Qwen3-30B, EP8   & 8 req/s   & 4.47 & 4.51 \\
                 & 16 req/s  & 4.45 & 4.50 \\
                 & 32 req/s  & 4.72 & 4.82 \\
                 & 64 conc.  & 5.21 & 5.28 \\
                 & 128 conc. & 6.54 & 6.46 \\
OLMoE, EP4       & 32 req/s  & 2.41 & 2.39 \\
                 & 64 conc.  & 2.83 & 2.78 \\
                 & 192 conc. & 4.00 & 3.90 \\
\bottomrule
\end{tabular}
\end{table}

\subsection{Every GPU Holding Every Expert}
\label{app:tp}

Table~\ref{tab:tp} builds up the MoE layer's speedup over vLLM's Triton MoE when every GPU holds
every expert, one change at a time. The kernel accounts for the gain, as Section~\ref{sec:eval-source}
expects when the batch size largely predicts the best configuration.

\begin{table}[h!]
\caption{\textbf{Every GPU holding every expert.} MoE layer speedup from vLLM's Triton MoE to \sys, one change per
column, over all batches. TP$n$: each expert split over $n$ GPUs.}
\label{tab:tp}
\vskip 0.08in
\centering
\footnotesize
\setlength{\tabcolsep}{3.5pt}
\begin{tabular}{lcccc}
\toprule
Model & Kernel & Trace-tuned & \sys & Total \\
\midrule
Qwen3-30B TP1 & 1.10 & 1.00 & 1.00 & 1.10 \\
Qwen3-30B TP2       & 1.16 & 1.01 & 1.00 & 1.17 \\
Qwen3.5-35B TP1 & 1.07 & 1.02 & 1.00 & 1.09 \\
Qwen3.5-35B TP2     & 1.16 & 1.01 & 0.98 & 1.14 \\
DeepSeek-V3 TP8     & 1.07 & 1.03 & 0.99 & 1.09 \\
\bottomrule
\end{tabular}
\end{table}

\subsection{The Polymorphic Megakernel}
\label{app:kernel}

\sys's kernel is written in CuTe DSL~\citep{cutlass} for Hopper and runs both expert projections of
a GPU's experts in one launch.

\textbf{Operand layout.} Both projections use FP8 WGMMA instructions with FP32 accumulation. The
expert's weights are the A operand, 64 output columns per instruction, and a tile's tokens are the
B operand, so the tokens of a tile form the instruction's N dimension. Both operands are K-major in
128-byte-swizzled shared memory, and one weight layout serves both projections. A thread block
covers 256 or 512 output columns with one producer warpgroup, which loads operands, and several
consumer warpgroups, which compute; \texttt{setmaxnreg} moves registers from the producer to the
consumers.

\textbf{Fusion.} Each tile runs in three phases. In the up-projection, the producer gathers the
tile's token rows in the layout the selection kernel wrote and loads the interleaved gate and up
weights with TMA, and the consumers multiply and apply the $128 \times 128$ block scales. In
registers, the consumers then compute the gated activation, quantize it to FP8 with a per-token
scale that also folds in the router's weight, and store it to shared memory. In the
down-projection, TMA loads the down-projection weights, the consumers multiply the FP8
intermediate, and the BF16 result is added into the layer's output with bulk asynchronous
reductions.

\textbf{Polymorphism.} Every menu entry is compiled into the same kernel, which branches on the
index the selection kernel writes. Entries share the thread-block size and may differ in rows per
tile, pipeline depth, tile width, persistence, and launch order; the served menus use one tile
width. The grid covers every entry's tiles, and blocks outside the chosen entry's grid exit at
once. All entries share one shared-memory buffer, sized for the largest entry, and one register count
that covers every entry (Table~\ref{tab:kernelres}).

\begin{table}[h!]
\caption{\textbf{Resources of the polymorphic kernel}, OLMoE. Shared memory and registers per thread
of each entry compiled alone, and of every entry inside the polymorphic kernel.}
\label{tab:kernelres}
\vskip 0.08in
\centering
\footnotesize
\setlength{\tabcolsep}{6pt}
\begin{tabular}{lcc}
\toprule
Menu & Entries alone: shared memory (KB) / registers & Inside the kernel \\
\midrule
3 entries & 85--103 / 72--126 & 103 / 126 \\
6 entries & 53--103 / 72--126 & 103 / 128 \\
\bottomrule
\end{tabular}
\end{table}

\end{document}